\documentclass[sigconf]{acmart}
\usepackage{subcaption}
\usepackage{bigstrut}
\usepackage{multirow}
\usepackage{colortbl}
\usepackage{xcolor}
\usepackage{pifont}
\usepackage{etoolbox}
\makeatletter
\patchcmd{\authornote}{\g@addto@macro\addresses{\@authornotemark}}{}{}{}
\makeatother

\AtBeginDocument{%
  }

\acmSubmissionID{4688}

\copyrightyear{2025} \acmYear{2025} \setcopyright{acmlicensed}\acmConference[MM '25]{Proceedings of the 33rd ACM International Conference on Multimedia}{October 27--31, 2025}{Dublin, Ireland} \acmBooktitle{Proceedings of the 33rd ACM International Conference on Multimedia (MM '25), October 27--31, 2025, Dublin, Ireland} \acmDOI{10.1145/3746027.3755600} \acmISBN{979-8-4007-2035-2/2025/10}

\begin{document}

\title{GalaxAlign: Mimicking Citizen Scientists' Multimodal Guidance for Galaxy Morphology Analysis}






\author{Ruoqi Wang}
\orcid{0009-0005-3513-1945}
\affiliation{%
  \institution{The Hong Kong University of Science and Technology (Guangzhou)}
  \city{Guangzhou}
  \country{China}}
\email{rwang280@connect.hkust-gz.edu.cn}

\author{Haitao Wang}
\orcid{0000-0002-8394-6410}
\affiliation{%
  \institution{School of Computer Science and Engineering, Sun Yat-Sen University}
  \city{Guangzhou}
  \country{China}
}
\email{wanght39@mail2.sysu.edu.cn}

\author{Qiong Luo}
\authornote{Qiong Luo is the corresponding author.}
\orcid{0000-0002-2861-9492}
\affiliation{%
  \institution{The Hong Kong University of Science and Technology (Guangzhou)}
  \city{Guangzhou}
  \country{China}}
\affiliation{%
 \institution{The Hong Kong University of Science and Technology}
 \city{Hong Kong}
 \country{China}}
 \email{luo@ust.hk}






\begin{abstract}
Galaxy morphology analysis involves studying galaxies based on their shapes and structures. For such studies, fundamental tasks include identifying and classifying galaxies in astronomical images, as well as retrieving visually or structurally similar galaxies through similarity search. Existing methods either directly train domain-specific foundation models on large, annotated datasets or fine-tune vision foundation models on a smaller set of images. The former is effective but costly, while the latter is more resource-efficient but often yields lower accuracy. To address these challenges, we introduce GalaxAlign, a multimodal approach inspired by how citizen scientists identify galaxies in astronomical images by following textual descriptions and matching schematic symbols. Specifically, GalaxAlign employs a tri-modal alignment framework to align three types of data during fine-tuning: (1) schematic symbols representing galaxy shapes and structures, (2) textual labels for these symbols, and (3) galaxy images. By incorporating multimodal instructions, GalaxAlign eliminates the need for expensive pretraining and enhances the effectiveness of fine-tuning.  Experiments on galaxy classification and similarity search demonstrate that our method effectively fine-tunes general pre-trained models for astronomical tasks by incorporating domain-specific multi-modal knowledge. Code is available at https://github.com/RapidsAtHKUST/GalaxAlign.
\end{abstract}




\maketitle

\section{Introduction}
\begin{figure}[ht]
    \begin{subfigure}[t]{0.9\linewidth}
        \includegraphics[width=\linewidth]{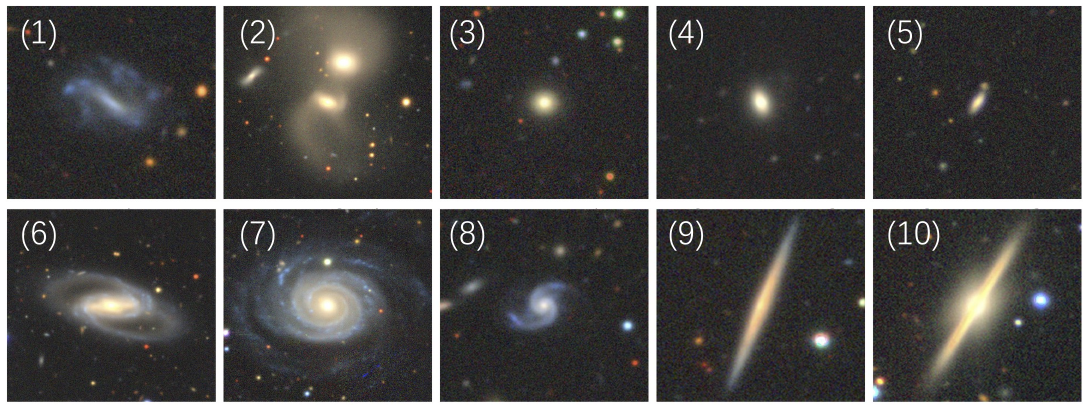}
        \caption{Examples of galaxy images.}\label{img:a2}
    \end{subfigure}

    \begin{subfigure}[t]{0.9\linewidth}
        \includegraphics[width=\linewidth]{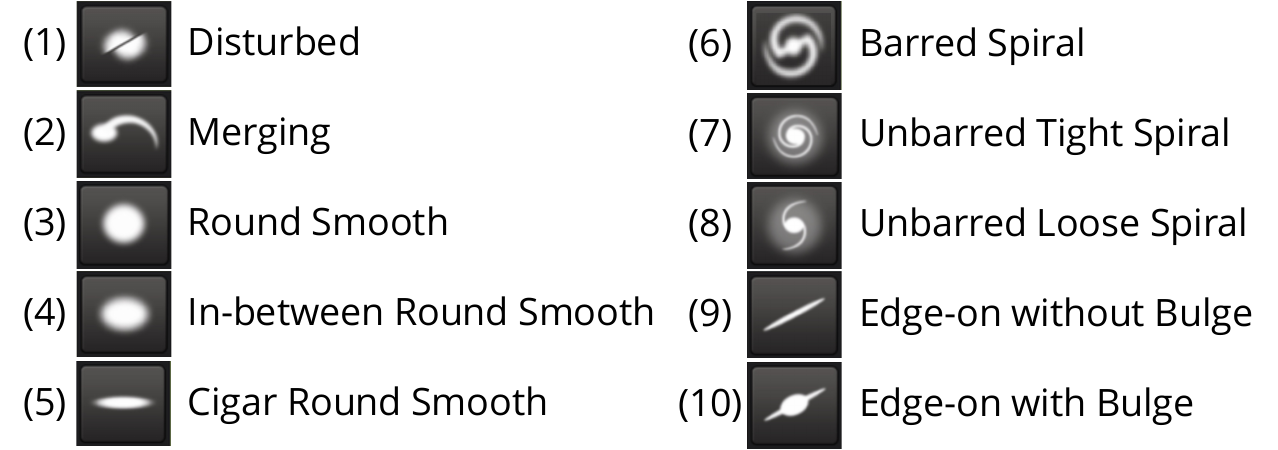}
        \caption{Examples of galaxy morphology description and schematic symbols corresponding to the images in subfigure (a).}\label{img:b2}
    \end{subfigure}\bigskip
\caption{Examples of galaxy images, corresponding schematic symbols and their textual labels. Images and textual labels are from the Galaxy10 DECaLS dataset \cite{galaxy10}, whereas schematic symbols are from the Galaxy Zoo 2 decision tree \cite{dieleman2015rotation, willett2013galaxy}.}
\label{egs}
\end{figure}

Galaxy morphology analysis involves studying galaxies based on their shapes and structures. This information is crucial for understanding galaxy formation and evolution, and it can be conveyed through natural language and schematic diagrams of galaxy images\cite{walmsley2020galaxy, lintott2011galaxy}. As shown in Figure \ref{egs}, schematic symbols paired with textual labels effectively capture the distinct characteristics of individual galaxies\cite{bowles2022new, bowles2023radio}. These textual descriptions and schematic diagrams have proven useful in instructing citizen scientists in galaxy image annotation\cite{walmsley2020galaxy}, where the annotations from citizen scientists serve as a critical pipeline for processing massive astronomical datasets. In this paper, we explore how to train an AI-based citizen scientist for galaxy image annotation, demonstrating the use of multi-modal instructions to enhance existing foundation models for galaxy morphology analysis. 

Foundation models pre-trained on large-scale natural image datasets, such as ImageNet \cite{russakovsky2015imagenet}, perform well in various domain applications. However, researchers commonly believe that these pre-trained models are insufficient for astronomical images \cite{walmsley2022practical, slijepcevic2024radio, lastufka2024vision}. This concern arises because the domain-specific scientific data differ significantly from the general natural image datasets used for foundation models, leading to distribution shifts that reduce the effectiveness of these models when applied directly to galaxy analysis \cite{lastufka2024vision}.

As a result, astronomical foundation models have typically relied on pretraining with large-scale astronomical datasets \cite{lintott2011galaxy, willett2013galaxy, dieleman2015rotation, lanusse2023astroclip}. These models are trained from scratch using extensive domain-specific datasets and subsequently fine-tuned for downstream tasks,  overlooking the potential benefits of adapting publicly available vision foundation models that are pre-trained on natural datasets. Moreover, annotating galaxy images heavily depends on the efforts of experts and volunteers, requiring significant human resources and time \cite{jimenez2023czsl, walmsley2020galaxy, lanusse2023astroclip}. Therefore, there is a pressing need for a method that reduces reliance on large domain-specific datasets and directly utilizes readily available pre-trained models trained on natural image data.

To address this need, we introduce \textbf{GalaxAlign}, a tri-modal framework that adapts pre-trained CLIP models \cite{radford2021learning} for galaxy morphology analysis by integrating schematic symbols, textual descriptions, and galaxy images. Since human amateur volunteers can label astronomical images based on their knowledge learned from textual descriptions and schematic diagrams, we believe that these modalities can also assist models pre-trained on general datasets in performing classification tasks on astronomical images.

Specifically, GalaxAlign employs a two-stage fine-tuning approach: In the first stage, galaxy images and schematic symbols are input into a shared image encoder, while textual descriptions are processed by a separate text encoder. This stage enables the image encoder to learn a shared representation of galaxy features from both symbolic and photographic images. 
In the second stage, GalaxAlign transfers the parameters from the shared encoder in Stage 1 to initialize a separate symbol encoder, enabling each encoder to specialize in a single modality—images, symbols, or text. This parameter transfer leverages the shared encoder’s foundational understanding of galaxy morphology, learned from both images and schematic symbols in Stage 1, as a strong starting point for Stage 2. The second stage fine-tuning aligns each modality for more precise feature embedding by focusing each encoder on its unique input.

This tri-modal alignment enhances the model’s ability to distinguish detailed structural features, improving classification and similarity search accuracy without costly large-scale pretraining. Extensive experiments show that GalaxAlign effectively fine-tunes pre-trained models, achieving high accuracy in astronomical tasks by incorporating domain-specific multi-modal knowledge.

Our contributions are as follows:
\begin{itemize}
    \item We propose \textbf{GalaxAlign}, the first tri-modal alignment framework that integrates galaxy images, schematic symbols, and natural language descriptions, effectively mimicking the cognitive strategy of citizen scientists in galaxy morphology analysis.
    \item We design a \textbf{two-stage multimodal adaptation pipeline} that first unifies visual-symbolic representations through a shared encoder, and then transitions to modality-specific encoders for fine-grained alignment, enabling efficient and accurate transfer from general vision-language models to domain-specific astronomical tasks.
\end{itemize}

\section{Background and Related Work} 
\label{sec:related}

\subsection{Foundation Models in Astrophysics}
Unlike natural images, astrophysics images have the following properties \cite{lastufka2024vision}: 
\begin{itemize}
    \item \textbf{Sparseness}: Objects occupy only a small fraction of each image.
    \item \textbf{Noise}: Systematic noise is present in the images.
    \item \textbf{High Dynamic Range}: Object brightness spans several orders of magnitude.
    \item \textbf{Artifacts}: Instrumental effects or reconstruction residuals introduce unintended structures of various scales.
\end{itemize}

Due to these differences, astronomers regard general vision models trained on natural images as inadequate for astronomical tasks. As such, instead of fine-tuning existing vision foundation models to fit astronomical data, Walmsley et al. proposed a galaxy morphology foundation model pre-trained on large-scale galaxy morphology datasets \cite{walmsley2022practical}. However, constructing such datasets is time-consuming and labor-intensive. For instance, the GZD-5 project, which classified 262,000 galaxies, spanned over three years from March 2017 to October 2020 \cite{walmsley2022galaxy}. Similarly, in the recent Galaxy Zoo campaign, 38,949 volunteers annotated a total of 105,000 galaxies over nearly two years (November 2020 to October 2022) \cite{walmsley2023galaxy}. 

To reduce the amount of time and labor required for creating large, labeled astronomical datasets, an alternative approach is to adapt existing foundation models pre-trained on natural images to astronomical tasks. Some representative pre-trained general vision foundation models including DINOv2 \cite{oquab2023dinov2}, MAE \cite{he2022masked},  MSN \cite{assran2022masked}, supervised ResNet \cite{he2016deep}, and supervised ViT \cite{dosovitskiy2020image}, have been fine-tuned to adapt to galaxy morphology tasks. However, their performance is generally poor \cite{lastufka2024vision}. These results suggest that adapting vision foundation models for astrophysics applications requires further considerations of data characteristics and alignment with domain-specific knowledge. Our work contributes to this effort.

\begin{figure}[t]
\centerline{\includegraphics[width=7.5cm]{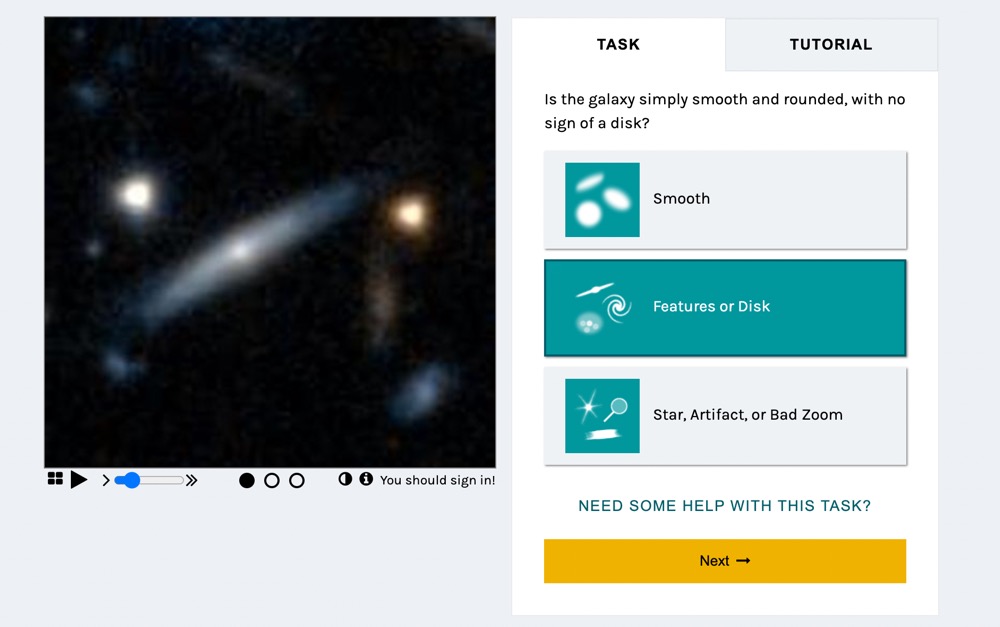}}
 \caption{Multi-modal annotation instructions for volunteers \cite{GZAnnotation}.}
\label{gz}
\end{figure}

\subsection{Citizen Science and Multi-Modal Data}
Citizen science projects are the primary method for large-scale galaxy morphology annotation. In the Galaxy Zoo projects \cite{lintott2011galaxy, willett2013galaxy, walmsley2022galaxy}, volunteers annotate astronomical images based on specific instructions. This process heavily relies on providing amateur volunteers with simplified schematic symbols and natural language descriptions to guide them through the classification process. 

Figure \ref{gz} illustrates an example of the annotation guidance available to volunteers on the Galaxy Zoo online platform \cite{GZAnnotation}. Volunteers are presented with symbols that depict different morphological features, such as spiral arms, bars, or mergers, accompanied by concise textual descriptions of these features. In this annotation process, the combination of visual symbols and natural language descriptions helps non-experts connect their common sense with astronomical knowledge, enabling them to effectively complete labeling tasks. 

Inspired by this approach, our method integrates schematic symbols and natural language descriptions as key components in galaxy morphology tasks, aiming to bridge the gap between pre-trained general models and astronomical data.

\subsection{Multi-Modal Contrastive Learning Models}
Vision-language models, such as CLIP \cite{radford2021learning}, utilize contrastive learning to align visual and textual embeddings. These models obtain strong generalization capabilities by training on large-scale datasets that pair images with corresponding textual descriptions. Bowles et al. \cite{bowles2022new, bowles2023radio} proposed a method to connect radio galaxy images with human-generated natural language descriptions to derive semantic morphology classes for classification, demonstrating that textual descriptions aligns with distinct identifying features in the images.

Moreover, multi-modal contrastive learning has been widely applied in various scientific domains \cite{liu2023text, imam2024cosmoclip, sanchez2023cloome, mishra2024paperclip, lanusse2023astroclip, vivanco2024geoclip}, learning semantic representations effectively across diverse modalities. In this paper, we present the first application of associating astronomical data with three modalities—images, text, and schematic symbols—demonstrating that contrastive learning can effectively align these modalities.

\section{Method}
\label{sec:method}

\begin{figure}
\centerline{\includegraphics[width=8cm]{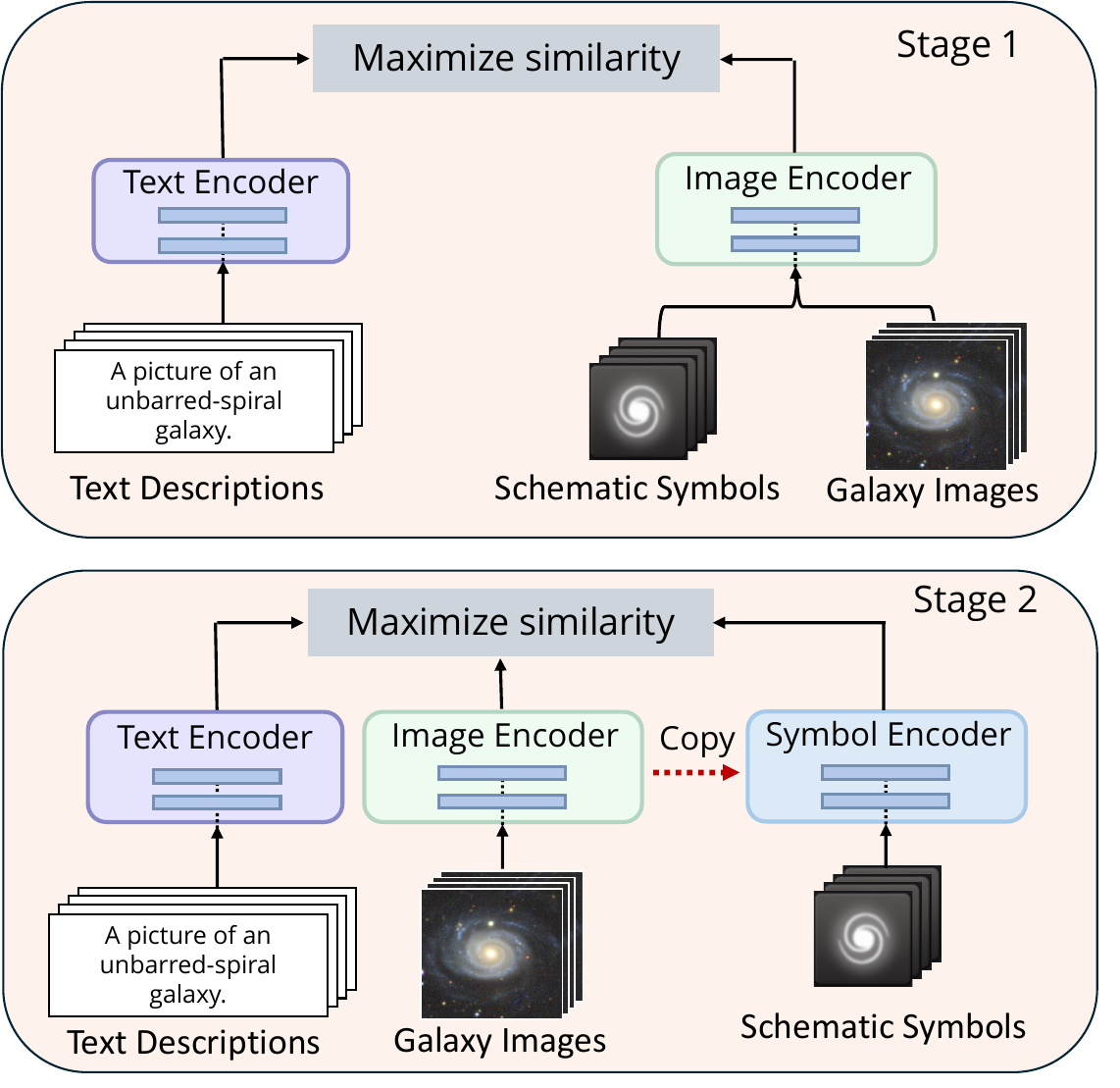}}
 \caption{Overview of our GalaxAlign. Stage 1 is the finetuning with schematic symbols and images sharing a single encoder. Stage 2 is the finetuning with 3 separate encoders. }
\label{Visual}
\end{figure}

We introduce GalaxAlign, a tri-modal learning framework based on the CLIP architecture, for galaxy morphology tasks by incorporating and aligning three modalities: textual descriptions, astronomical images, and schematic symbols. 

Our framework contains three encoders:
\begin{itemize}
    \item \textbf{Text Encoder:} It processes textual descriptions related to galaxy morphology.
    \item \textbf{Image Encoder:} Initially it encodes both galaxy images and schematic symbols to enable the learning of a shared representation of the two modalities. After the parameters from the shared representation are learned, the image encoder encodes galaxy images only.
    \item \textbf{Symbol Encoder:} It is derived from the Image Encoder after the first stage to encode schematic symbols in the second stage.
\end{itemize}

Our finetuning is conducted in two stages: 
\begin{itemize}
    \item \textbf{Stage 1:} Symbol-image with text training.
    \item \textbf{Stage 2:} Tri-modal joint training. 
\end{itemize}

\subsection{Textual Description}
Our input text follows the common format \textit{A picture of a/an \{class name\}}. Since the \textit{class name} clearly and concisely describes the structure of the galaxies, it provides a succinct and distinguishing textual label. For example, sentences such as “\textit{A picture of an unbarred-spiral galaxy}” or “\textit{A picture of a cigar round smooth galaxy}” effectively capture the structural information of the galaxies.

\subsection{Stage 1: Symbol-Image with Text Training}
In the first stage, the Image Encoder takes both galaxy images and symbols as its input, whereas the Text Encoder processes corresponding text descriptions. This stage aligns the representations of images and symbols with text descriptions, setting up a shared multi-modal embedding space. The shared Image Encoder extracts patterns in galaxy shapes and structures that are consistent across symbols and images. Both encoders are initialized with pre-trained weights of the CLIP model \cite {ilharco_gabriel_2021_5143773} on natural images.

For each input galaxy image \( x_{\text{img}} \), schematic symbol \( x_{\text{sym}} \), and text description \( x_{\text{txt}} \), we obtain embeddings:
\[
z_{\text{img}} = E_{\text{img}}(x_{\text{img}}), \quad z_{\text{sym}} = E_{\text{img}}(x_{\text{sym}}), \quad z_{\text{txt}} = E_{\text{txt}}(x_{\text{txt}})
\]

To align image/symbol and text embeddings, we use a contrastive loss function that maximizes the cosine similarity between positive (matching) pairs and minimizes it for non-matching pairs. For batch size \( N \), the loss function for image-text and symbol-text pairs is:
\[
\mathcal{L}_{\text{con}} = -\frac{1}{2N} \sum_{i=1}^{N} \left( \log \frac{\exp(\text{sim}(z_{\text{img/sym}}^i, z_{\text{txt}}^i)/\tau)}{\sum_{j=1}^{N} \exp(\text{sim}(z_{\text{img/sym}}^i, z_{\text{txt}}^j)/\tau)} \right)
\]

where \( \text{sim}(z_a, z_b) \) denotes the cosine similarity and \( \tau \) is a learnable temperature parameter.

In our method, \textbf{Stage 1 serves as a warm-up phase} rather than full- convergence training. Empirically, we found that training for just over 10 epochs in our experiments was sufficient to achieve optimal results, eliminating the need for full convergence at this stage. This approach allows the model to establish a solid foundation for galaxy morphology analysis without excessive computation in the initial stage.

\subsection{Stage 2: Tri-Modal Joint Training}
In Stage 2, we transition from a shared encoder to separate, modality-specific encoders, allowing the encoders to refine and specialize in the unique features of images, symbols, and text individually. To facilitate this specialization, we copy the parameters from the shared image encoder \( E_{\text{img}} \) from Stage 1 to initialize the symbol encoder \( E_{\text{sym}} \):
\[
E_{\text{sym}} \leftarrow E_{\text{img}}
\]
With the symbol encoder starting from the same representation as the image encoder, both encoders are more likely to produce embeddings that are well-aligned in the feature space. Moreover, rather than learning from scratch, the symbol encoder refines and specializes existing representations from Stage 1, allowing itself to capture modality-specific details more quickly.

All three encoders are then fine-tuned together, optimizing the alignment across text, image, and symbol representations. The resulting embeddings are:
\[
z_{\text{img}} = E_{\text{img}}(x_{\text{img}}), \quad z_{\text{sym}} = E_{\text{sym}}(x_{\text{sym}}), \quad z_{\text{txt}} = E_{\text{txt}}(x_{\text{txt}})
\]

The Stage 2 loss function includes three contrastive components to ensure effective alignment across all three modalities. We define the modality pairs as:
\[
(a, b) \in \{(\text{img}, \text{txt}), (\text{img}, \text{sym}), (\text{sym}, \text{txt})\}
\]

For each pair, the contrastive loss is computed as:

\[
\mathcal{L}_{\text{total}} = -\frac{1}{N} \sum_{i=1}^{N} \sum_{(a, b)} \log \frac{\exp(\text{sim}(z_{a}^i, z_{b}^i)/\tau)}{\sum_{j=1}^{N} \exp(\text{sim}(z_{a}^i, z_{b}^j)/\tau)}
\]

This loss function facilitates a balanced alignment across the three modalities, enabling the model to jointly learn textual descriptions, visual images, and schematic symbols for galaxy morphology tasks.

This two-stage process enables the model to adapt pretrained terrestrial models to the specialized tasks of galaxy morphology classification, retrieval, and similarity search, utilizing both textual descriptions and schematic visual symbols as complementary modalities. Our experiments include comparisons with the state-of-the-art pretrained models under full-data, few-shot and zero-shot settings, demonstrating the effectiveness of our multimodal strategy in fine-tuning general models without the need for large-scale astronomical pretraining.

\section{Experiments}
In this section, we first present the visualization of galaxy feature embeddings from our model in comparison with a number of existing methods. Then, we conduct comparative studies on the accuracy performance of two tasks - galaxy morphology classification and similarity.  Finally, we perform few-shot tests and ablation studies. More details regarding the implementation, datasets and additional experimental results are reported in the appendix. 

\subsection{Experimental Setup}

\subsubsection{Platform}
We conduct all experiments on a server with two AMD EPYC 7543 CPUs, 512GB main memory, and four NVIDIA RTX A6000 GPUs each with 48GB device memory. The operating system is Ubuntu 22.04. Our model is implemented in PyTorch 2.1.0.

\subsubsection{Datasets}
In our experiments, we evaluate our method on two representative public galaxy datasets: (1) Galaxy10 DECaLS \cite{galaxy10} and (2) GalaxyMNIST \cite{GalaxyMNIST}. Galaxy10 contains 17,736 colored galaxy images divided in 10 classes, with each image of a size $256 \times 256$ pixels.  This dataset is from the DESI Legacy Imaging Surveys \cite{dey2019overview}, which merges data from the Beijing-Arizona Sky Survey (BASS) \cite{zou2017project}, the DECam Legacy Survey (DECaLS) \cite{blum2016decam}, and the Mayall z-band Legacy Survey \cite{silva2016mayall}. In comparison, GalaxyMNIST \cite{GalaxyMNIST}, derived from Galaxy Zoo DECaLS \cite{walmsley2022galaxy}, contains 10,000 galaxy images ($64 \times 64$) of four morphological classes. 

\subsubsection{Evaluation Metrics}
In this study, we evaluate the performance of GalaxAlign on classification tasks using Accuracy and F1 Score (macro), providing a general sense of the model's classification capability across the dataset as well as offering a balanced view of performance across all classes. 
For the similarity search task, we use mean Average Precision (mAP), which reflects the ranking quality by calculating the average precision at various recall levels for each query and then averaging across all queries. 


\begin{figure*}
    \includegraphics[width=17cm]{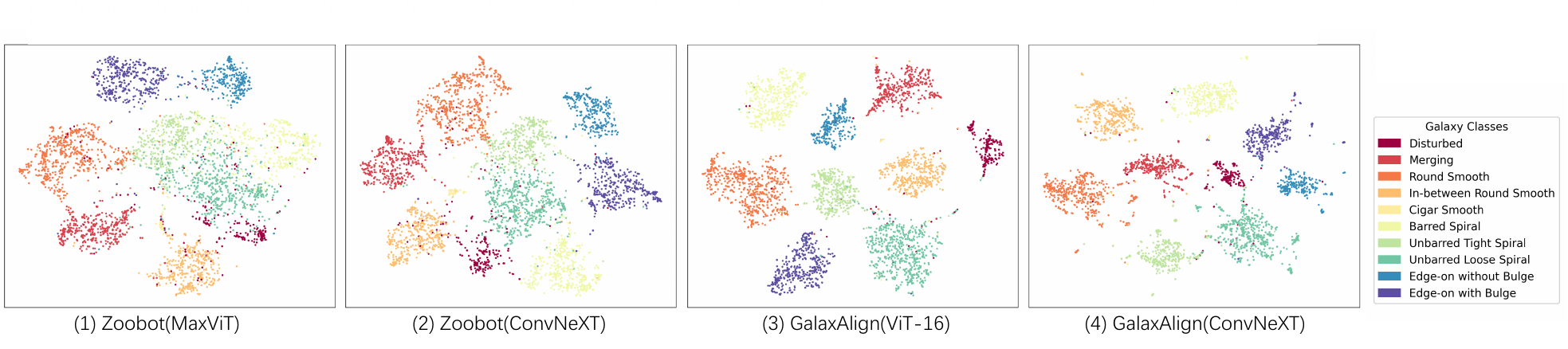} 
    \caption{The t-SNE visualization for features extracted using Zoobot (MaxViT and ConvNeXT) and GalaxAlign (ViT-16 and ConvNeXT) on Galaxy10 dataset.}
    \label{tsne}
\end{figure*}

\begin{figure*}
\centerline{\includegraphics[width=10cm]{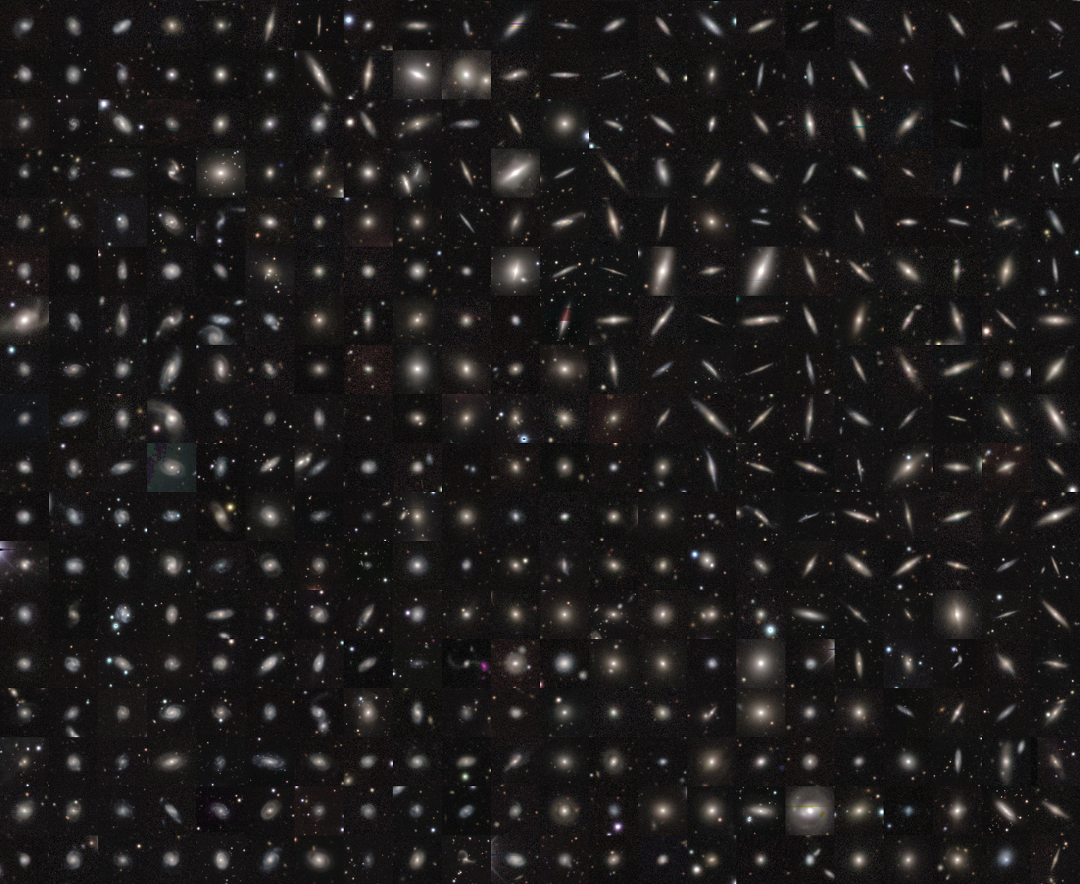}}
 \caption{Visualisation of the representations learned by our method, illustrating similar galaxies occupying nearby regions in the feature space. This visualization is created using PCA to compress the representation to 2D and placing galaxy thumbnails at the locations of their corresponding galaxies in the grid.}
\label{grid_visualization}
\end{figure*}

\subsubsection{Methods under Comparison}

We evaluate GalaxAlign with state-of-the-art foundation models, including mainstream general vision models, vision-language models and specialized astronomical foundation models trained on large-scale domain-specific datasets. The baseline models include:

\textbf{MAE} \cite{he2022masked}, \textbf{DINOv2} \cite{oquab2023dinov2}, \textbf{MSN} \cite{assran2022masked}: Vision foundation models using self-supervised pretraining techniques. Following Lastufka et al. \cite{lastufka2024vision}, we fine-tune the models using galaxy datasets of variant sizes in downstream tasks.

\textbf{ViT-16} \cite{dosovitskiy2020image}, \textbf{ResNet-50} and \textbf{ResNet-18} \cite{he2016deep}: Standard vision models pretrained on ImageNet-1k \cite{he2016deep} with a supervised method for classification.

\textbf{CLIP} \cite{radford2021learning}: Standard vision-language models pretrained on Data-Comp-1B \cite{gadre2023datacomp} and Laion-2b \cite{schuhmann2022laion}.
    
\textbf{Zoobot} \cite{walmsley2023zoobot}: State-of-the-art astronomical models, pretrained specifically on large-scale galaxy data from scratch. We selected MaxVIT and ConvNeXT as the backbone networks for Zoobot because they performed the best among all backbones \cite{walmsley2023zoobot}.

\begin{table}[h]
\centering
\caption{Models used in our comparison study.}
\label{tab:comparison_models}
\small
\begin{tabular}{l|l|l}
\hline
\textbf{Model Name} & \textbf{Backbone} &  \textbf{Pre-training Dataset} \\ \hline
MAE \cite{he2022masked}  & ViT-Base/16   & ImageNet-1k \cite{deng2009imagenet}     \\ \hline
DINOv2 \cite{oquab2023dinov2}  & ViT-Base/14   & LVD-142M \cite{oquab2023dinov2}    \\ \hline
MSN \cite{assran2022masked}  & ViT-Base/16   & ImageNet-1k \cite{deng2009imagenet}    \\ \hline
ViT-16 \cite{dosovitskiy2020image}  & ViT-Base/16   & ImageNet-1k \cite{deng2009imagenet}    \\ \hline
ResNet 50 \cite{he2016deep}  & ResNet 50   & ImageNet-1k \cite{deng2009imagenet}    \\ \hline
ResNet 18 \cite{he2016deep}  & ResNet 18   & ImageNet-1k \cite{deng2009imagenet}    \\ \hline
\multirow{2}{*}{CLIP \cite{radford2021learning}}   & ViT-16-Base    & DataComp-1B \cite{gadre2023datacomp}  \\ \cline{2-3} 
& ConvNext-Base    & LAION-2B \cite{schuhmann2022laion}   \\ \hline
\multirow{2}{*}{Zoobot \cite{walmsley2023zoobot}}   & MaxViT-Base    & GZ DECaLS GZD-5 \cite{walmsley2022galaxy}   \\ \cline{2-3} 
& ConvNeXT-Base   & GZ DECaLS GZD-5 \cite{walmsley2022galaxy}    \\ \hline
\end{tabular}
\end{table}

\subsection{Feature Projections}
To demonstrate GalaxAlign's strong performance without relying on extensive domain-specific datasets, we present the embedding visualization comparing our method with Zoobot, which has been pretrained on large astronomical datasets and then fine-tuned on smaller datasets. Figure \ref{tsne} provides a t-SNE visualization \cite{van2008visualizing} of galaxy data embeddings of Galaxy10 dataset learned by different models. Feature embeddings of other baseline methods and datasets are presented in the appendix.

In Figure \ref{tsne} (a), both Zoobot models (MaxViT and ConvNeXT) display notable overlaps between complex classes including Barred Spiral Galaxy, Unbarred Tight Spiral and Unbarred Loose Spiral, indicating limited performance in distinguishing fine morphological details. In contrast, our GalaxAlign models (ViT-16 and ConvNeXT) exhibit well-separated, clearly defined clusters with minimal overlap, demonstrating effectiveness in category distinction and feature representation. Meanwhile, GalaxAlign consistently maintains distinct clusters across all categories, reflecting its ability to capture essential morphological features accurately. In Figure \ref{tsne} (b), Zoobot models exhibit some clustering but display considerable overlap in the Smooth Cigar and Edge-on Disk categories. In contrast, GalaxAlign models achieve more distinct, well-separated clusters across all categories. Overall, these results highlight GalaxAlign’s advantage over Zoobot in achieving compact intra-class clustering and distinct inter-class separability, underscoring its effectiveness in representing galaxy morphology.

Figure \ref{grid_visualization} shows a 2D grid visualization of galaxy images sampled from the GalaxyMNIST dataset, where high-dimensional features are reduced using PCA and aligned to a grid. The clusters and smooth transitions between similar galaxy types in the grid suggest that our model effectively encodes structural information in the feature space, allowing visually similar galaxies to be placed in close proximity. The visualization of images in the Galaxy10 dataset is presented in the appendix.

\begin{table*}[h]
\centering
\caption{Comparison of Classification Performance (mean and standard deviation) on GalaxyMNIST and Galaxy10}
\label{tab:performance_comparison}
\small
\begin{tabular}{l|c|c|c|c|c|c}
\toprule
\multirow{2}{*}{\textbf{Method}} & \multicolumn{2}{c|}{\textbf{Pretraining Dataset}} & \multicolumn{2}{c|}{\textbf{GalaxyMNIST}} & \multicolumn{2}{c}{\textbf{Galaxy10}} \\ \cline{2-7} 
                                & \textbf{General}     & \textbf{Domain}    & \textbf{Accuracy} & \textbf{F1 Score}    & \textbf{Accuracy} & \textbf{F1 Score}    \\ \hline

MAE & \ding{51} & & 0.7312 (0.0070) & 0.7314 (0.0069) & 0.6242 (0.0041) & 0.5990 (0.0098) \\ \hline
DINOv2 & \ding{51} & & 0.8786 (0.0013) & 0.8789 (0.0014) & 0.8465 (0.0008) & 0.8337 (0.0010) \\ \hline
MSN & \ding{51} & & 0.8275 (0.0016) & 0.8279 (0.0017) & 0.6113 (0.0030) & 0.5616 (0.0032) \\ \hline
ViT-16 & \ding{51} & & 0.8519 (0.0021) & 0.8521 (0.0021) & 0.7304 (0.0007) & 0.7054 (0.0015) \\ \hline
ResNet-18  & \ding{51} &   &  0.8720 (0.0150) & 0.8722 (0.0151) &  0.9501 (0.0123) & 0.9446 (0.0117) \\ \hline
ResNet-50   & \ding{51} &  & 0.8877 (0.0059) & 0.8884 (0.0059) & 0.9466 (0.0023) & 0.9399 (0.0040)  \\ \hline
CLIP (ViT-16)  & \ding{51} &   &  0.9125 (0.0006) & 0.9129 (0.0005) &  0.9635 (0.0006) & 0.9549 (0.0007) \\ \hline
CLIP (ConvNeXT)  & \ding{51} &   &  0.9310 (0.0030) & 0.9316 (0.0028) &  0.9625 (0.0021) & 0.9537 (0.0018) \\ \hline
Zoobot (MaxViT) & & \ding{51} & 0.8790 (0.0022) & 0.8796 (0.0023) & 0.8922 (0.0065) & 0.8847 (0.0066) \\ \hline
Zoobot (ConvNeXT) & & \ding{51} & 0.9360 (0.0009) & 0.9365 (0.0009) & 0.9600 (0.0061) & 0.9550 (0.0062) \\ \hline
\rowcolor{gray!20} \textbf{Ours (ViT-16)} & \ding{51} & & 0.9272 (0.0005) & 0.9276 (0.0004) & \textbf{0.9732 (0.0004)} & \textbf{0.9702 (0.0005)} \\ \hline
\rowcolor{gray!20} \textbf{Ours (ConvNext)} & \ding{51} & & \textbf{0.9372 (0.0015)} & \textbf{0.9377 (0.0015)} & 0.9710 (0.0014) & 0.9664 (0.0008) \\ \bottomrule
\end{tabular}
\end{table*}

\begin{figure}
\centerline{\includegraphics[width=8.6cm]{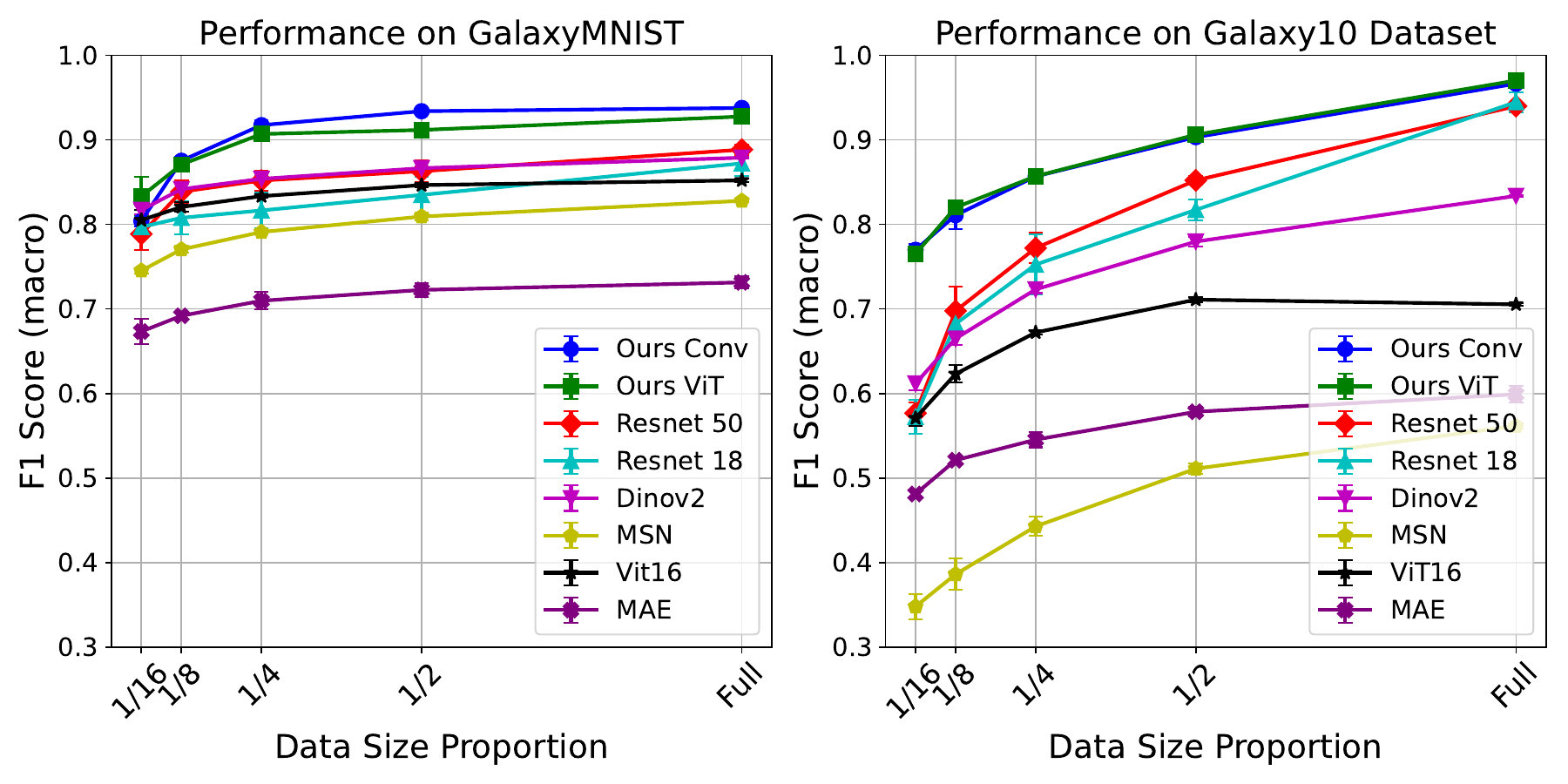}}
 \caption{Performance Comparison of Different Methods on GalaxyMNIST and Galaxy10 Datasets with Different Data Sizes (F1 Score).}
\label{line_datasize}
\end{figure}
\subsection{Galaxy Morphology Classification}
Table \ref{tab:performance_comparison} provides a comparative analysis of classification performance across different methods on the GalaxyMNIST and Galaxy10 datasets. 

The self-supervised models (MAE \cite{he2022masked}, DINOv2 \cite{oquab2023dinov2}, and MSN \cite{assran2022masked}), when fine-tuned following Lastufka et al. \cite{lastufka2024vision}, show promising performance on GalaxyMNIST but are generally outperformed by models pretrained with supervised learning \cite{dosovitskiy2020image, he2016deep}, particularly on Galaxy10. This result suggests that finetuning supervised models pretrained on natural images tends to transfer more effectively to the morphological distinctions in astronomical images. And fine-tuning CLIP \cite{radford2021learning} models with images and textual descriptions also achieve better performance than other uni-modal methods.

The Zoobot models \cite{walmsley2023zoobot} (using MaxViT and ConvNeXT as backbones, respectively), pretrained on large-scale galaxy data, demonstrate strong performance. This result underscores the value of domain-specific pretraining for astronomical classification.

In contrast, our proposed models—Ours (ViT-16) and Ours (ConvNext)—pretrained on natural image datasets and fine-tuned using a multi-modal architecture that aligns textual descriptions, schematic symbols, and astronomical images, achieve comparable performance to Zoobot and outperform all other methods. These results highlight the effectiveness of adapting general pretrained models through multi-modal fine-tuning, providing an alternative to large-scale astronomical pretraining for galaxy morphology tasks.

We also conduct additional evaluation using Gemini-2.5 and QVQ via the latest public APIs. We also finetuned LLaVA1.6 \cite{liu2023visual} using LoRA \cite{hu2022lora} on our full training datasets. The testing results on these baselines are presented in Appendix.

\begin{table}[h]
\centering
\caption{Comparison of Mean Average Precision (mAP) on GalaxyMNIST and Galaxy10.}
\label{mAP}
\small
\begin{tabular}{l|c|c|c|c}
\toprule
\multirow{2}{*}{\textbf{Method}} & \multicolumn{2}{c|}{\textbf{GalaxyMNIST}} & \multicolumn{2}{c}{\textbf{Galaxy10}}  \\ \cline{2-5}
 & \textbf{mAP@5} & \textbf{mAP} & \textbf{ mAP@5} & \textbf{mAP} \\ \hline
MAE & 0.5867 & 0.2571 & 0.5592 & 0.1302 \\ \hline
DINOv2 & 0.6472 & 0.2566 & 0.6193 & 0.1261 \\ \hline
MSN & 0.6372 & 0.3005 & 0.5569 & 0.1508 \\ \hline
ViT-16 & 0.7929 & 0.4208 & 0.6800 & 0.2229 \\ \hline
ResNet-18 & 0.9160 & 0.6865 & 0.9398 & 0.5827 \\ \hline
ResNet-50 & 0.9184 & 0.6657 & 0.8732 & 0.5431 \\ \hline
Zoobot (MaxViT) & 0.9011 & 0.8408 & 0.9370 & 0.7842 \\ \hline
Zoobot (ConvNeXT) & 0.9406 & 0.9123 & 0.9491 & 0.8492 \\ \hline
\rowcolor{gray!20} Ours (ViT-16) & 0.9524 & 0.8865 & \textbf{0.9919} & \textbf{0.9645} \\ \hline
\rowcolor{gray!20} Ours (ConvNeXT) & \textbf{0.9569} & \textbf{0.9123} & 0.9868 & 0.9640 \\ \bottomrule
\end{tabular}
\end{table}

\begin{figure}[h]
    \begin{subfigure}[t]{1\linewidth}
        \includegraphics[width=\linewidth]{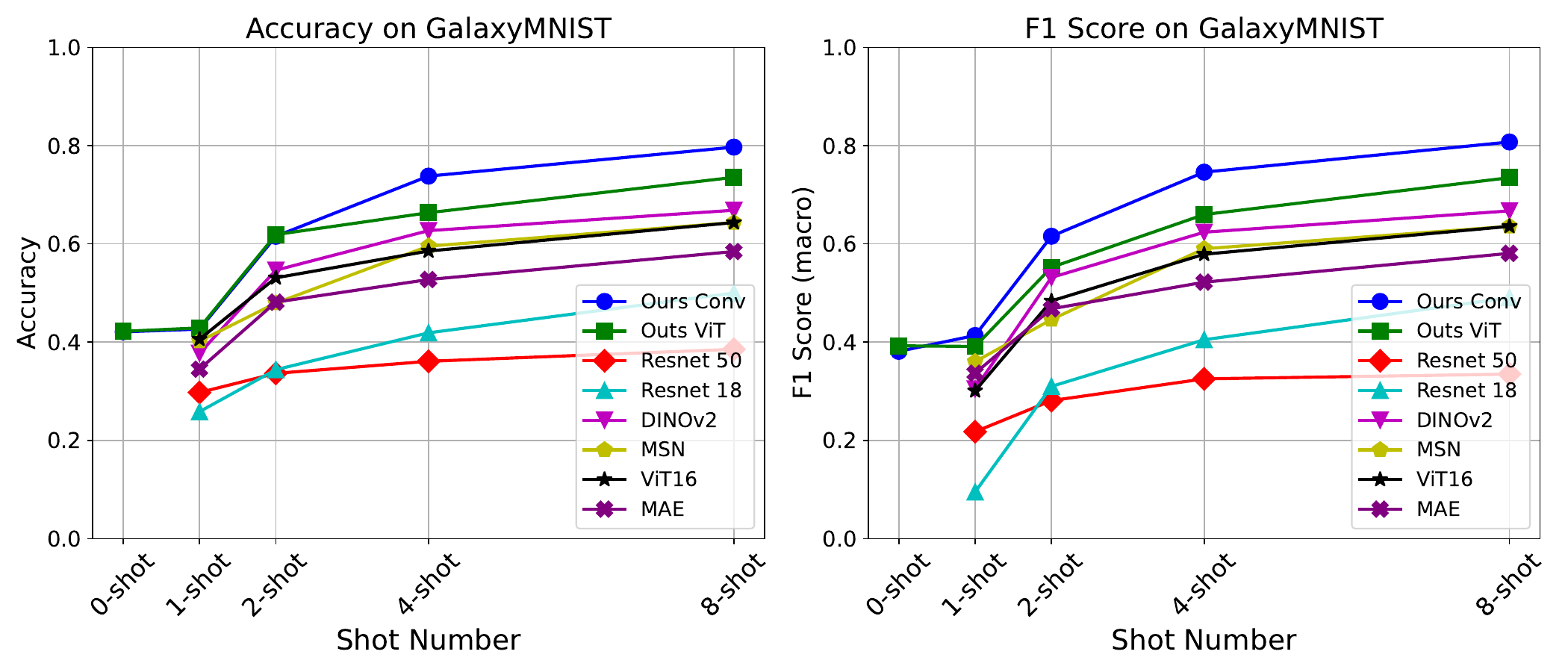}
        \caption{Few-shot tests on GalaxyMNIST dataset.}\label{img:a}
    \end{subfigure}

    \begin{subfigure}[t]{1\linewidth}
        \includegraphics[width=\linewidth]{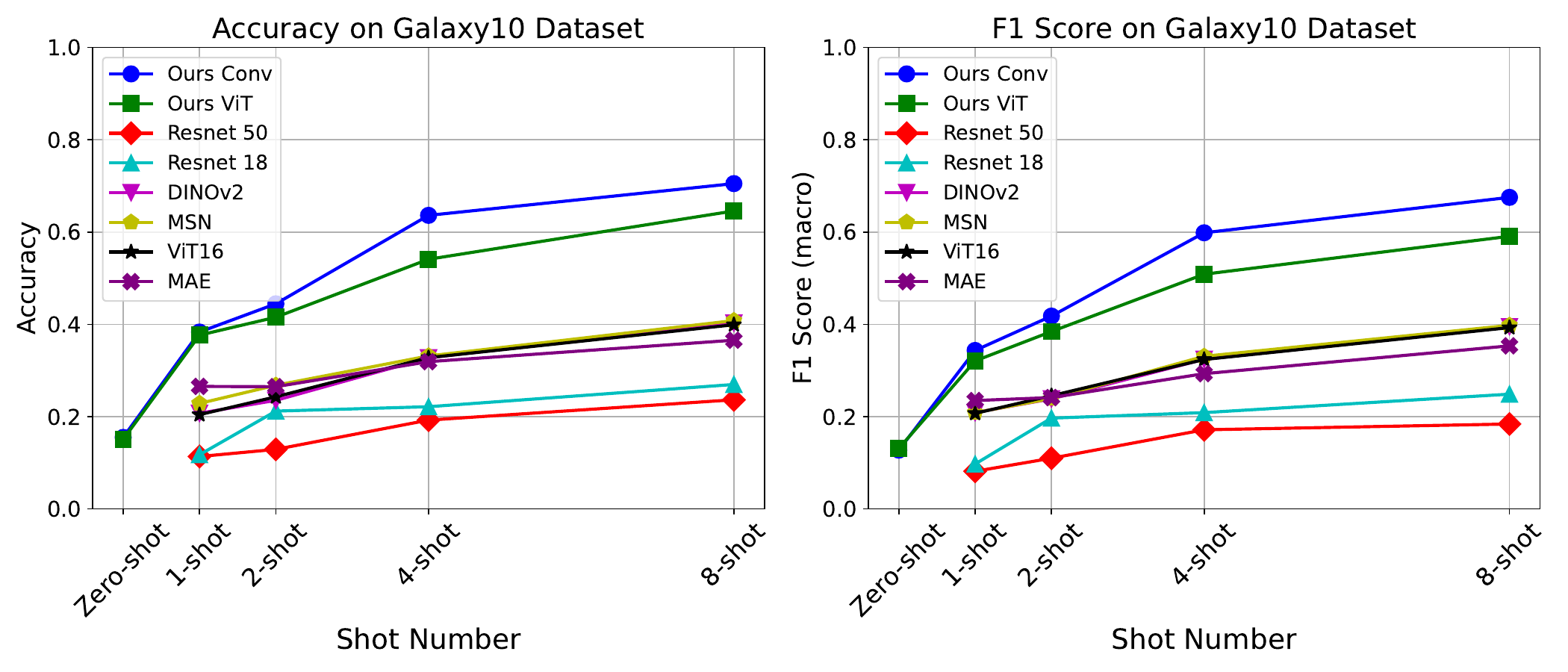}
        \caption{Few-shot tests on Galaxy10 dataset.}\label{img:b}
    \end{subfigure}\bigskip
\caption{Few-Shot Tests of Different Methods on
GalaxyMNIST and Galaxy10 Datasets.}
\label{fewshot}
\end{figure}

\begin{figure*}[h]
\centerline{\includegraphics[width=15.5cm]{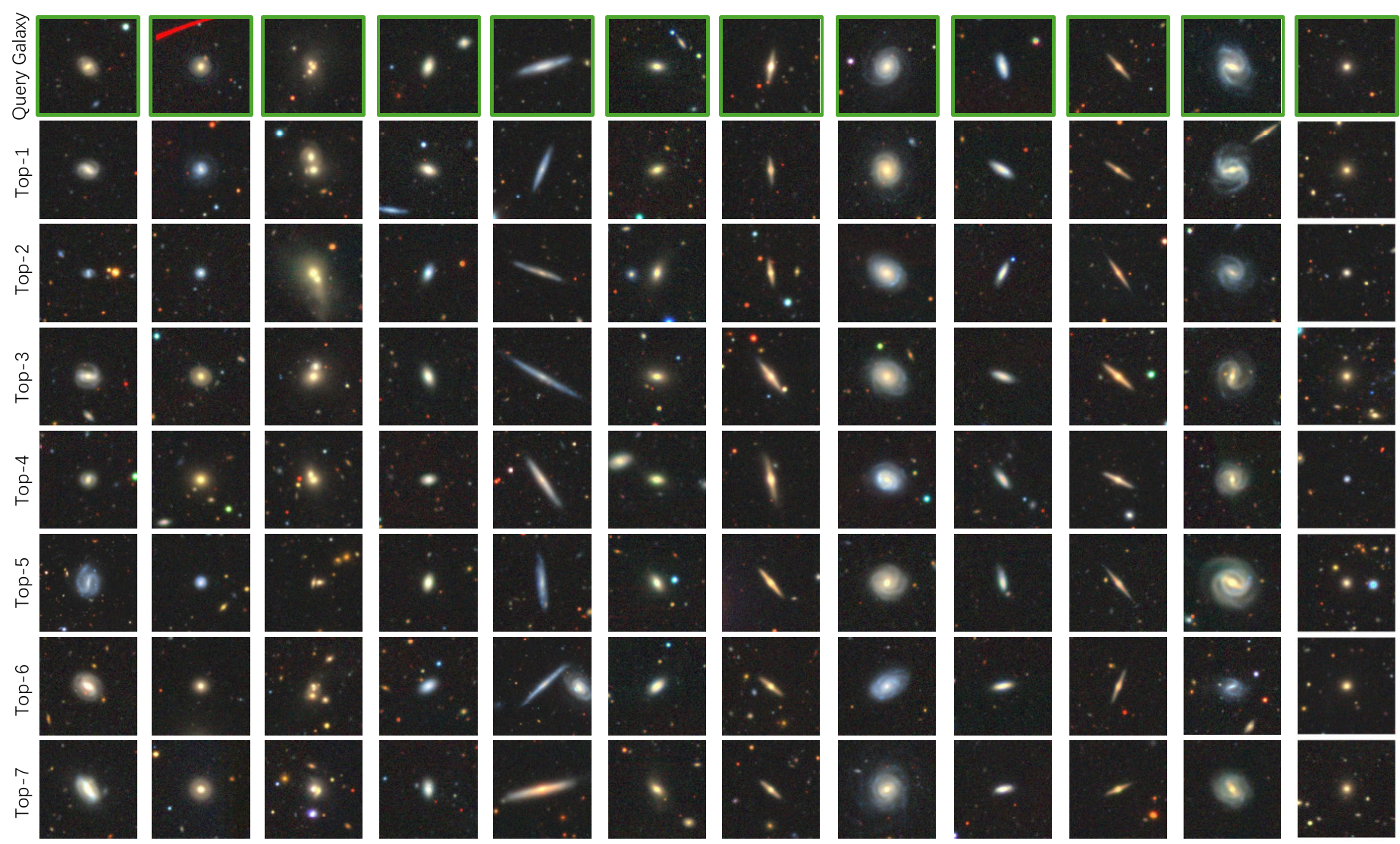}}
 \caption{Examples of similarity search on the Galaxy10 dataset. The top row shows the query galaxy images (outlined in green) for each column, and the other rows display the seven most similar galaxies retrieved by the model. }
\label{similarity_search}
\end{figure*}

Figure \ref{line_datasize} compares the F1 scores of various models pretrained on natural data across different data sizes for the fine-tuning datasets, GalaxyMNIST and Galaxy10. Our models achieve the highest performance on both datasets, with all data sizes, showing efficient generalization and effective adaptation to astronomical tasks. Self-supervised models (e.g., DINOv2 \cite{oquab2023dinov2}, MAE \cite{he2022masked}, MSN \cite{assran2022masked}) and standard supervised models (e.g., ResNet-18, ResNet-50 \cite{he2016deep} and ViT \cite{dosovitskiy2020image}) achieve moderate results, indicating that pretraining with natural image alone may not fully capture the domain-specific details required for astronomical tasks. These results illustrate the effectiveness of our multi-modal adaptation strategy in bridging general pretrained models and domain-specific applications.


\subsection{Similarity Search}
In astronomy, automatically identifying the similarity between two galaxies is a challenging but essential task \cite{walmsley2022practical}. Effective searches for similar galaxies can help us find counterparts of rare galaxies, making the leap from a one-off discovery to a new class of phenomena \cite{walmsley2022practical}. Latent representations encoded by neural networks provide a new opportunity for measuring morphological similarity. In our similarity search evaluation, we measure the morphological similarity of galaxies by comparing feature embeddings through a similarity matrix. Specifically, we calculate pairwise similarities using dot products between feature vectors, creating a matrix where each entry indicates the similarity value between two galaxy images. For each image, we retrieve the most similar images (excluding the image itself) and use these to evaluate retrieval performance.

The comparison results of GalaxAlign and other methods evaluated in mAP@5 and mAP (mAP@all) are shown in Table \ref{mAP}. Our models demonstrate superior performance over baseline methods on both GalaxyMNIST and Galaxy10 datasets. While other fine-tuned general-purpose models show limited effectiveness, our approach, incorporating multi-modal alignment, outperforms traditional architectures, achieving better results in identifying galaxy morphological similarities. This result underscores our model’s capacity to adapt well to astronomical data.

The results in Figure \ref{similarity_search} demonstrate that our model effectively retrieves galaxies with similar morphological characteristics to the query. Across various galaxy types, the top-ranked matches closely resemble the query galaxy's shape and structure, indicating strong model performance in identifying morphological similarities.

\subsection{Few-Shot Tests}
We conduct few-shot tests on GalaxAlign and other foundation models pretrained on natural data. In Figure \ref{fewshot}, across all data sizes, our GalaxAlign model (labeled as "Ours") consistently achieves the highest or nearly the highest performance compared to other baseline models, highlighting GalaxAlign's strong adaptability to the specific characteristics of astronomical images. This advantage is particularly evident on the Galaxy10 dataset, where the increased number of morphological classes better demonstrates the effectiveness of our model.

\subsection{Ablation Studies}
We conduct an ablation study to evaluate the contribution of each alternative in our approach. First, we compare how well our model works with ViT and ConvNeXT. Then we examine the effect of our tri-modality input in comparison with the text-image bimodal CLIP model without the schematic symbol input. Finally, we compare our models across different stages of fine-tuning: Ours\_{v1} is the warm-up model trained through the first stage, omitting the second training stage; Ours\_{v2} skips the first stage and directly goes into the second stage of training until convergence; Ours\_{v3} omits the second stage, training through the first stage to convergence; Ours\_{Scratch} goes through both stages, but is trained from scratch on the Galaxy10 and GalaxyMNIST datasets, without pre-training on large-scale natural datasets. The results in Table \ref{tab:ablation_results} show the impact of each alternative in our approach. The full models, Ours(ViT-16) and Ours(ConvNeXT), achieve the highest accuracy and F1 scores on both datasets. Removing the symbol modality (CLIP models) or omitting training stages (e.g., Ours\_{v1}, Ours\_{v2}, and Ours\_{v3}) reduces performance. The Ours\_{Scratch} variant, trained without pretraining, shows notable drops, indicating the impact of pretraining with large-scale general datasets on astronomy tasks.

We also conduct the ablation study of text and symbols and the ablation study on each part of our tri-model loss function. Results in Table \ref{tab:ablation_results2} show that removing either the schematic symbols, text descriptions in our framework or each modal pair in the loss function leads to a performance drop. Each modality contributes to our tasks in a complementary way.

\begin{table}[h]
\setlength\tabcolsep{4pt}
\centering
\caption{Ablation Study of Training Porcess}
\label{tab:ablation_results}
\small
\begin{tabular}{l|cc|cc}
\toprule
\multirow{2}{*}{\textbf{Model Variant}} & \multicolumn{2}{c|}{\textbf{Galaxy10}} & \multicolumn{2}{c}{\textbf{GalaxyMNIST}} \\ \cline{2-5} 
 & \textbf{Accuracy} & \textbf{F1} & \textbf{Accuracy} & \textbf{F1} \\ \hline
CLIP(ViT-16) & 0.9635 & 0.9549 & 0.9125 & 0.9129 \\ \hline
Ours\_{v1} (ViT-16) & 0.9457 & 0.9398 & 0.9125 & 0.9132 \\ \hline
Ours\_{v2} (ViT-16) & 0.9647 & 0.9607 & 0.8965 & 0.8967 \\ \hline
Ours\_{v3} (ViT-16) & 0.9678 & 0.9624 & 0.9245 & 0.9247 \\ \hline
Ours\_{Scratch}(ViT-16) & 0.9585 & 0.9558 & 0.8620 & 0.8621 \\ \hline
\rowcolor{gray!20} Ours(ViT-16) & \textbf{0.9732} & \textbf{0.9702} & \textbf{0.9272} & \textbf{0.9276} \\ \bottomrule
\hline
CLIP(ConvNeXT) & 0.9625 & 0.9537 & 0.9310 & 0.9316 \\ \hline
Ours\_{v1} (ConvNeXT) & 0.9428 & 0.9330 & 0.8600 & 0.8596 \\ \hline
Ours\_{v2} (ConvNeXT) & 0.9656 & 0.9593 & 0.9260 & 0.9266 \\ \hline
Ours\_{v3} (ConvNeXT) & 0.9651 & 0.9607 & 0.9360 & 0.9365 \\ \hline
Ours\_{Scratch}(ConvNeXT) & 0.9599 & 0.9568 & 0.9301 & 0.9236 \\ \hline
\rowcolor{gray!20} Ours(ConvNeXT) & \textbf{0.9710} & \textbf{0.9664} & \textbf{0.9372} & \textbf{0.9377}  \\ \bottomrule
\end{tabular}
\end{table}

\begin{table}[h]
\setlength\tabcolsep{4pt}
\centering
\caption{Ablation Study of Modalities and Loss}
\label{tab:ablation_results2}
\small
\begin{tabular}{l|cc|cc}
\toprule
\multirow{2}{*}{\textbf{Model Variant}} & \multicolumn{2}{c|}{\textbf{Galaxy10}} & \multicolumn{2}{c}{\textbf{GalaxyMNIST}} \\ \cline{2-5} 
 & \textbf{Accuracy} & \textbf{F1} & \textbf{Accuracy} & \textbf{F1} \\ 
\midrule
\multicolumn{5}{l}{\textbf{ViT}} \\
\hline
w/o Symbol & 0.9635 & 0.9549 & 0.9125 & 0.9129 \\
w/o Text & 0.7965 & 0.7627 & 0.9050 & 0.9053 \\
Loss w/o text-image & 0.9617 & 0.9582 & 0.9100 & 0.9102 \\
Loss w/o symbol-image & 0.9622 & 0.9576 & 0.9125 & 0.9127 \\
Loss w/o text-symbol & 0.9634 & 0.9587 & 0.8945 & 0.9108 \\
\rowcolor{gray!20} Ours (Full) & \textbf{0.9732} & \textbf{0.9702} & \textbf{0.9272} & \textbf{0.9276} \\
\midrule
\multicolumn{5}{l}{\textbf{ConvNeXT}} \\
\hline
w/o Symbol & 0.9625 & 0.9537 & 0.9310 & 0.9316 \\
w/o Text & 0.7599 & 0.7330 & 0.8975 & 0.8952 \\
Loss w/o text-image & 0.5781 & 0.5503 & 0.7390 & 0.7379 \\
Loss w/o symbol-image & 0.9552 & 0.9444 & 0.9360 & 0.9364 \\
Loss w/o text-symbol & 0.9560 & 0.9491 & 0.9340 & 0.9346 \\
\rowcolor{gray!20} Ours (Full) & \textbf{0.9710} & \textbf{0.9664} & \textbf{0.9372} & \textbf{0.9377} \\
\bottomrule
\end{tabular}
\end{table}

\section{Conclusion and Future Work}
In this paper, we introduced GalaxAlign, a tri-modal framework designed to finetune vision foundation models for galaxy morphology analysis, utilizing schematic symbols, text descriptions, and galaxy images. By integrating domain-specific knowledge in a multi-modal approach, GalaxAlign effectively reuses the general foundation models pre-trained on natural datasets and reduces the need for manually labeled data to train foundation models from scratch. 

While GalaxAlign is designed for galaxy data, the multi-modal framework—integrating images, textual descriptions, and schematic symbols—is well-suited to other natural sciences such as biology for cellular and species classification, or geology for analyzing mineral structures, where both structure and descriptive information are essential and available. 

\section*{Acknowledgements}{
This work was supported by the China National Science Foundation (NSF) No. 62372393 and a startup fund from HKUST (Guangzhou).}

\bibliographystyle{ACM-Reference-Format}
\bibliography{sample-base}

\appendix

\clearpage
\setcounter{page}{1}


\twocolumn[{%
\renewcommand\twocolumn[1][]{#1}%
\begin{center}
\centering
\captionsetup{type=figure}
\includegraphics[width=17cm]{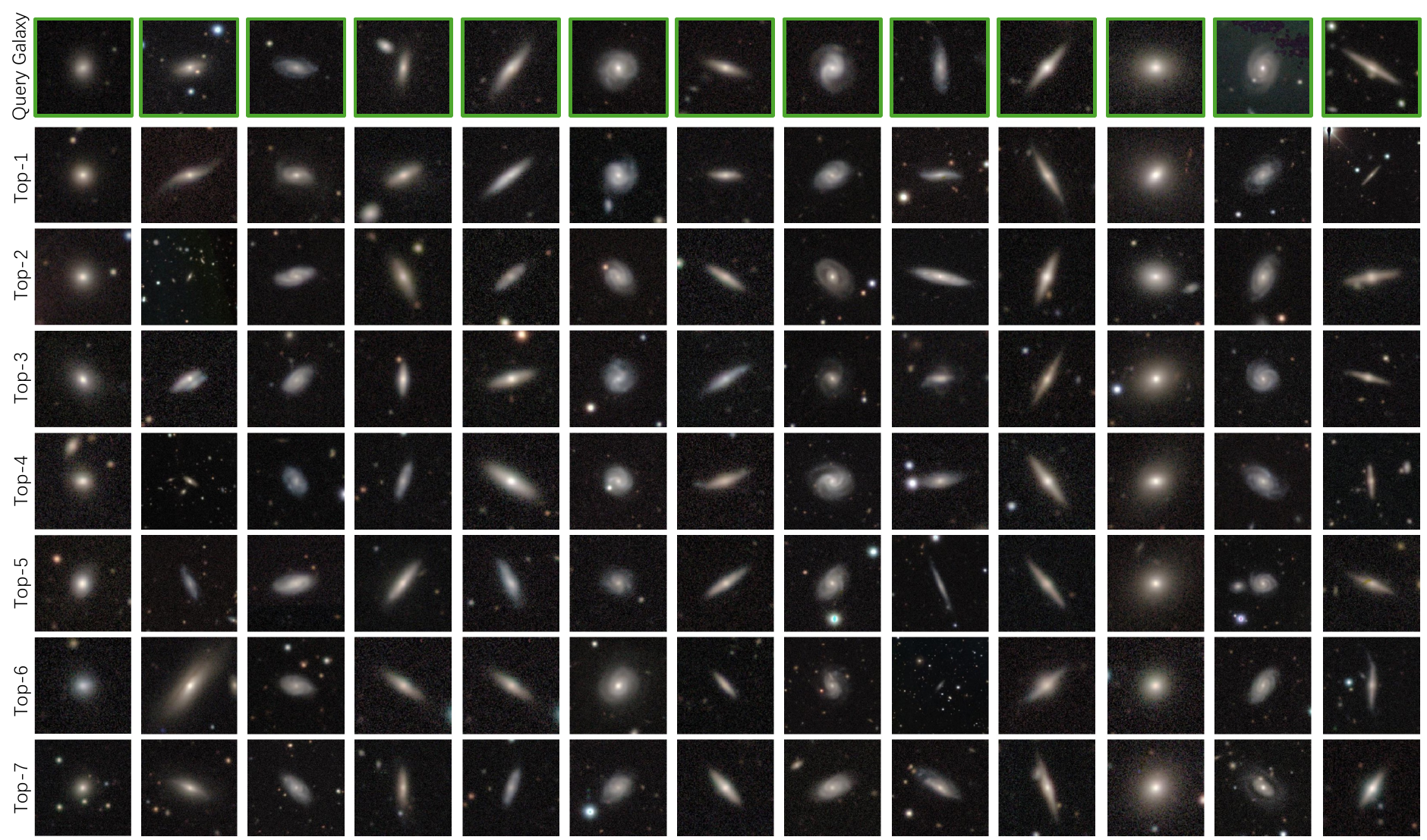}
 \caption{Examples of similarity search performed by GalaxAlign (ConvNeXT) on the GalaxyMNIST dataset. The top row shows the query galaxy images (outlined in green) for each column, and the other rows display the seven most similar galaxies retrieved by the model. }
\label{similarity_search_mnist}
\end{center}
}]

This document provides supplementary information that is not included in our main paper due to space limitation. We further
explain details about the experimental setup and present additional results. We provide our demo code in the supplementary material. All the code we used for the experiments will be public if the paper is published.

\setcounter{section}{0}
\renewcommand{\thesection}{\Alph{section}}

\section{Additional Experiment Details}
\subsection{Setup Details}
The ConvNeXT version of our model is trained with Adam with $lr = 5e-6, \text{weight decay} = 0.0002$. The ViT version of our model is trained with Adam with $lr = 1e-4, \text{weight decay} = 0.02$. We adopt a full fine-tuning strategy for our method. For the first stage, we train 50 epochs for the best results. For each dataset, we randomly split 20\% of all images for testing, with the remainder being used for training. We repeat all experiments five times and compute the mean and standard deviation of the results.

\subsection{Baseline Details}
To ensure a fair comparison and achieve the best performance for each baseline, we followed the fine-tuning strategies recommended in prior work. Specifically, for MAE \cite{he2022masked}, DINOv2 \cite{oquab2023dinov2}, MSN \cite{assran2022masked}, and ViT-16 \cite{dosovitskiy2020image}, we adopted the approach outlined by Lastufka et al. \cite{lastufka2024vision}, which involves freezing the encoder parameters and fine-tuning a classification head. In contrast, for ResNet18, ResNet50 \cite{he2016deep}, and Zoobot \cite{walmsley2023zoobot}, we used the same full fine-tuning strategy as our method, where all parameters in the model are updated during training, to maximize their performance on the tasks. This ensures that each baseline is optimized using its most suitable training procedure.

\section{Additional Results}
\subsection{Similarity Search}
Figure \ref{similarity_search_mnist} presents the results of a similarity search performed by GalaxAlign (ConvNeXT) on the GalaxyMNIST dataset. Each column represents a query galaxy (top row, outlined in green), followed by the seven most similar galaxies retrieved by the model (subsequent rows). The results demonstrate the model's ability to effectively retrieve galaxies with similar visual and morphological features. Across various galaxy types, the top-ranked matches closely resemble the query galaxy's shape and structure, indicating strong model performance in identifying morphological similarities.

\begin{figure*}
\centerline{\includegraphics[width=14cm]{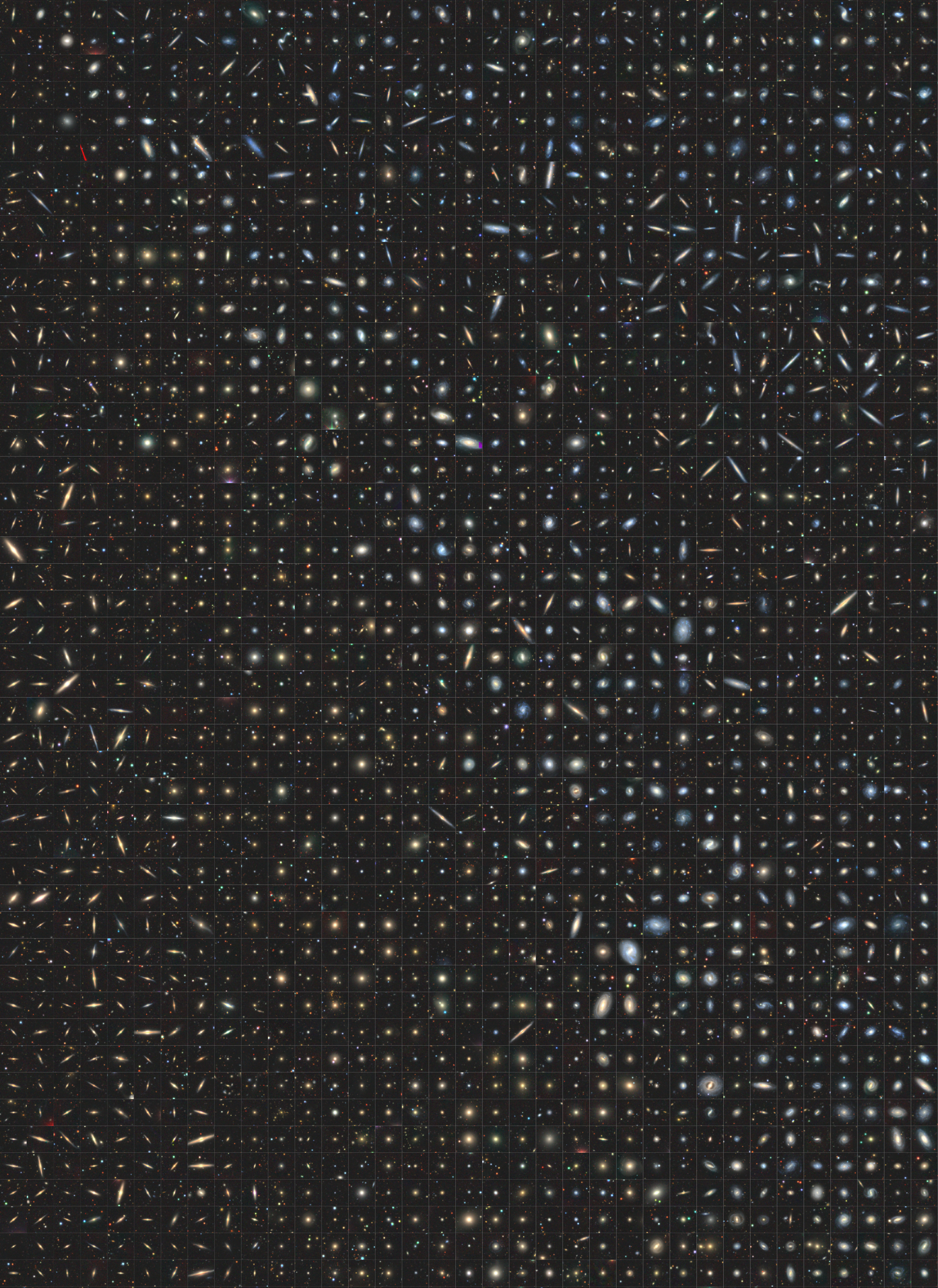}}
 \caption{Visualization of the representations learned by our method on the Galaxy10 dataset, illustrating similar galaxies occupying nearby regions in the feature space. This visualization is created using PCA to compress the representation to 2D and place galaxy images at the locations of their corresponding galaxies in the grid.}
\label{grid_visualization_ga10}
\end{figure*}

\begin{figure}
\centerline{\includegraphics[width=8cm]{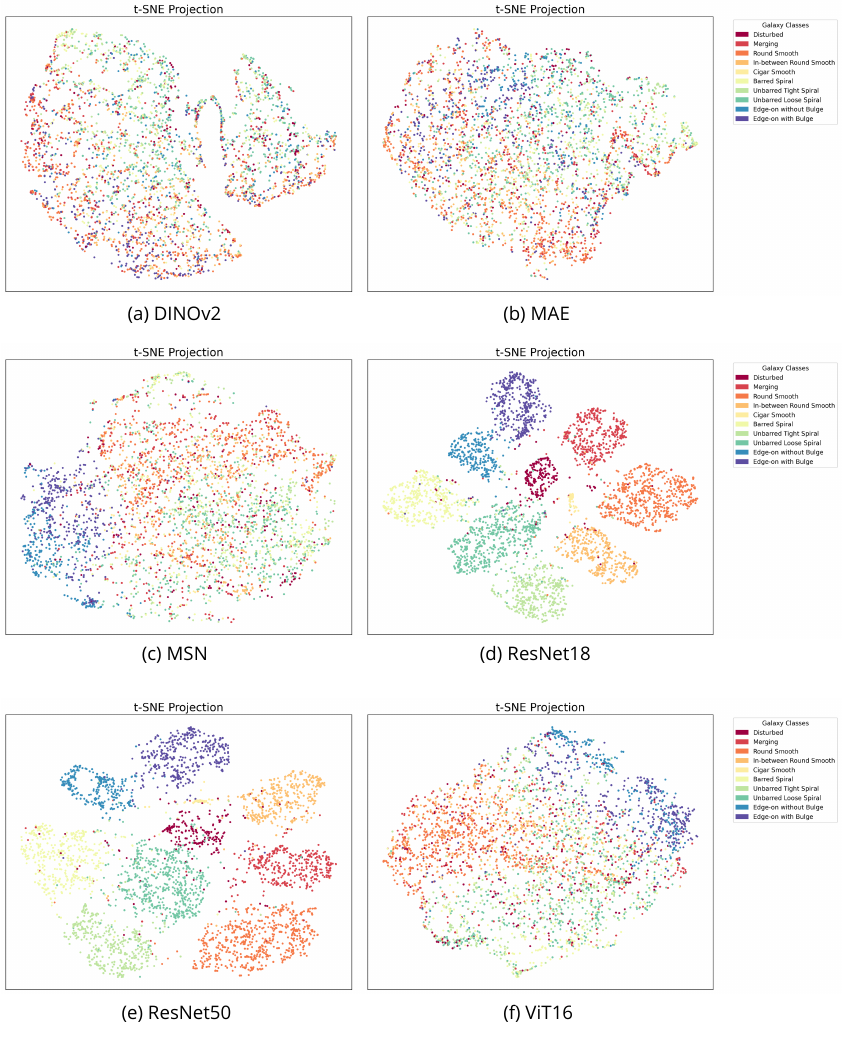}}
 \caption{The t-SNE visualization for features extracted using DINOv2, MAE, MSN, ResNet18, ResNet50, ViT on Galaxy10 dataset.}
\label{others_ga10}
\end{figure}

\begin{figure}
\centerline{\includegraphics[width=8cm]
{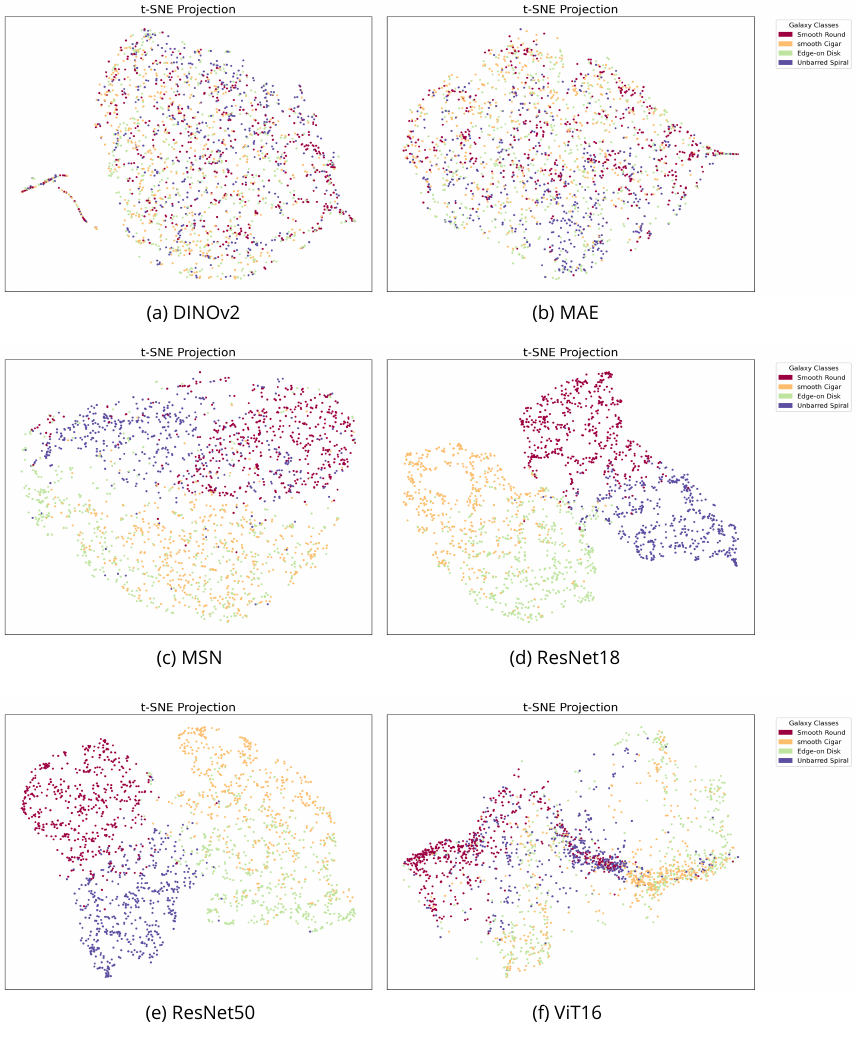}}
 \caption{The t-SNE visualization for features extracted using DINOv2, MAE, MSN, ResNet18, ResNet50, ViT on GalaxyMNIST dataset.}
\label{others_mnist}
\end{figure}

\subsection{Additional Comparison with Large Models}
We did additional evaluation using Gemini-2.5 and QVQ via the latest public APIs. We also finetuned LLaVA1.6 using LoRA on our full training datasets. The testing results on these baselines are in Table \ref{tab:model_comparisonadd}. GalaxAlign consistently outperforms these methods, highlighting the importance of our tri-modal approach.

\begin{table}[h]
\setlength\tabcolsep{6pt}
\centering
\caption{Additional Performance Comparison on Galaxy10 and GalaxyMNIST}
\label{tab:model_comparisonadd}
\small
\begin{tabular}{l|cc|cc}
\toprule
\textbf{Model} & \multicolumn{2}{c|}{\textbf{Galaxy10}} & \multicolumn{2}{c}{\textbf{GalaxyMNIST}} \\
\cline{2-5}
 & \textbf{Acc} & \textbf{F1} & \textbf{Acc} & \textbf{F1} \\
\midrule
Gemini-2.5        & 0.375 & 0.239 & 0.550 & 0.534 \\
QVQ               & 0.160 & 0.170 & 0.425 & 0.440 \\
LLaVA1.6 (LoRA)   & \textbf{0.917} & \textbf{0.905} & \textbf{0.896} & \textbf{0.898} \\
\bottomrule
\end{tabular}
\end{table}

\subsection{Additional Feature Projections}
\begin{figure*}
    \centering
    \includegraphics[width=1\linewidth]{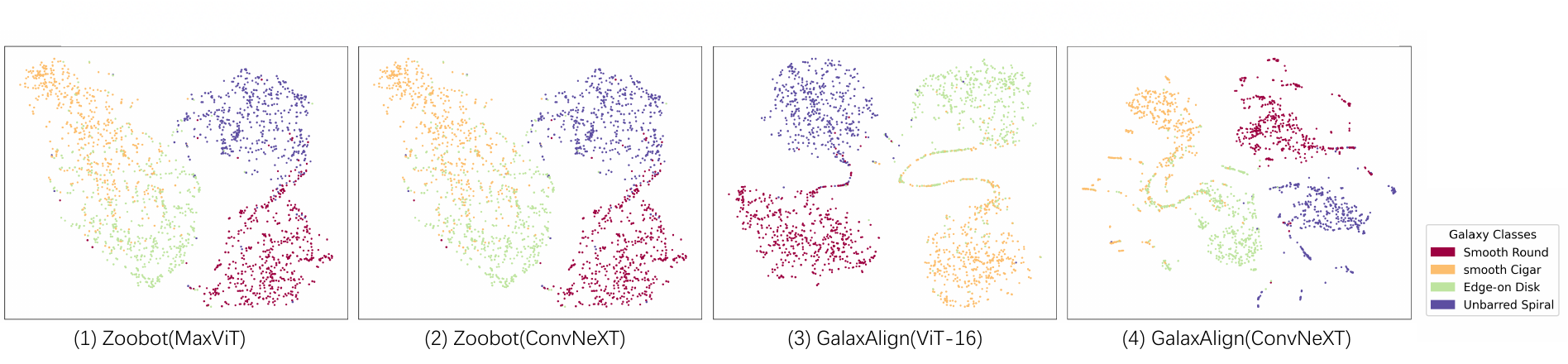}
    \caption{The t-SNE visualization for features extracted using Zoobot (MaxViT and ConvNeXT) and GalaxAlign (ViT-16 and ConvNeXT) GalaxyMNIST dataset.}
    \label{mnistvis}
\end{figure*}

Figure \ref{mnistvis} provides a t-SNE visualization \cite{van2008visualizing} of galaxy data embeddings of GalaxyMNIST dataset learned by different models.
Figure \ref{others_ga10} and Figure \ref{others_mnist} illustrate the t-SNE visualizations of features extracted from various baseline models, including DINOv2, MAE, MSN, ResNet18, ResNet50, and ViT16, on the Galaxy10 and GalaxyMNIST datasets, respectively. These visualizations demonstrate the separability of different galaxy classes in the feature space, providing insights into how well the models capture morphological differences among galaxy types.

For the Galaxy10 dataset, ResNet50 and ResNet18 exhibit moderately distinct clusters, indicating their ability to differentiate between galaxy classes. However, the clusters remain partially overlapping, particularly for morphologically similar classes (e.g., Barred Spiral, Unbarred Tight Spiral and Unbarred Loose Spiral). Models such as DINOv2, MAE and MSN show significant class mixing, reflecting suboptimal feature representations. ViT16 demonstrates slightly better clustering compared to DINOv2, MAE and MSN but still struggles to achieve complete separability.

On the GalaxyMNIST dataset, the separability of galaxy classes improves slightly for most models. ResNet18 and ResNet50 again outperform other baselines, showing relatively distinct clusters, although some class overlaps persist. Other models produce dispersed embeddings, with notable mixing between classes, highlighting their limitations in capturing galaxy morphology.

Figure \ref{grid_visualization_ga10} provides a 2D grid visualization of galaxy images sampled from the Galaxy10 dataset. Similar to the GalaxyMNIST visualization, high-dimensional features were reduced using PCA and aligned to a grid. The arrangement highlights the structural similarity between galaxies, with morphologically similar types clustering together in adjacent regions. The smooth transitions between different galaxy classes further demonstrate the ability of our model to effectively capture and encode galaxy morphology in the feature space, enabling clear distinctions and meaningful groupings across diverse galaxy types.

\end{document}